# Synthesis of neural networks for spatio-temporal spike pattern recognition and processing

**Authors:** J. Tapson[1*], G. Cohen[1], S. Afshar[1], K. Stiefel[1], Y. Buskila[1], R. Wang[1], T.J. Hamilton[1], A. van Schaik[1]

[1]The MARCS Institute, University of Western Sydney, Kingswood, NSW, Australia

**Correspondence:**

Prof Jonathan Tapson
School of Computing, Engineering and Mathematics
University of Western Sydney
Locked Bag 1797
Penrith 2751 NSW
Australia
j.tapson@uws.edu.au

**Abstract:**

The advent of large scale neural computational platforms has highlighted the lack of algorithms for synthesis of neural structures to perform predefined cognitive tasks. The Neural Engineering Framework offers one such synthesis, but it is most effective for a spike rate representation of neural information, and it requires a large number of neurons to implement simple functions. We describe a neural network synthesis method that generates synaptic connectivity for neurons which process time-encoded neural signals, and which makes very sparse use of neurons. The method allows the user to specify – arbitrarily - neuronal characteristics such as axonal and dendritic delays, and synaptic transfer functions, and then solves for the optimal input-output relationship using computed dendritic weights. The method may be used for batch or online learning and has an extremely fast optimization process. We demonstrate its use in generating a network to recognize speech which is sparsely encoded as spike times.

1. Introduction

There has been significant research over the past two decades to develop hardware platforms which are optimized for spiking neural computation. These platforms range from analog VLSI systems in which neurons are directly simulated by using CMOS transistors as ion channels and synapses, to highly parallel custom silicon microprocessor arrays (Boahen, 2006; Khan *et al*., 2008; Schemmel *et al*., 2010).

Some of these platforms are now capable of modeling populations of over a million neurons, at rates which are significantly faster than biological real time.

The advent of these systems has revealed a lack of concomitant progress in algorithmic development, and particularly in the synthesis of spiking neural networks. While there are a number of canonical structures, such as Winner-Take-All (WTA) networks (Indiveri, 2001), and some spiking visual processing structures such as Gabor filter networks and convolutional neural networks are routinely implemented (Zamarreño-Ramos *et al*., 2013), there are few successful methods for direct synthesis of networks to perform any arbitrary task which may be defined in terms of spike inputs and spike outputs, or in terms of a functional input-output relationship.

One successful method is the Neural Engineering Framework (NEF) (Eliasmith and Anderson, 2003). The NEF was first described in 2003 and generally makes use of a standard three-layer neural structure, in which the first layer are inputs; the second layer is a very large hidden layer of nonlinear interneurons, which may have recurrent connections; and the third layer is the output layer, which consists of neurons with linear input-output characteristics. The connections between the input and hidden layers are randomly weighted, and fixed (they are not altered during training). The connections between the hidden and output layers are trained in a single pass, by mathematical computation rather than incremental learning. We will describe this structure in more detail in the following section.

The NEF was perhaps the first example of a larger class of networks which have been named LSHDI networks – Linear Solutions of Higher Dimensional Interlayers (Tapson and van Schaik, 2013). These are now widely used in the machine learning community in the form of the Extreme Learning Machine (ELM) (Huang *et al*., 2006) – a conventional numerical neural network, which performs with similar accuracy to Support Vector Machines (SVMs) and which is significantly quicker to train than SVMs. Both the ELM and NEF methods have been applied to implement bio-inspired networks on neural computation hardware (Galluppi *et al*., 2012; Choudhary *et al*., 2012; Conradt *et al*., 2012; Basu *et al*., 2012). Most recently, Eliasmith and colleagues have used the method as the basis for a 2.5 million neuron simulation of the brain (Eliasmith *et al*., 2012).

The NEF is an effective synthesis method, with three important caveats: it intrinsically uses a spike rate-encoded information paradigm; it requires a very large number of neurons for fairly simple functions (for example, it is not unusual for a function with two inputs and one output, to use an interlayer of fifty to a hundred spiking neurons); and the synthesis (training) of weights is by mathematical computation using a singular value decomposition, rather than by any biologically plausible learning process.

We have recently addressed the third of these caveats by introducing weight synthesis in LSHDI through an online, biologically plausible learning method called OPIUM - the Online PseudoInverse Update Method (Tapson and van Schaik, 2013). This method also allows for adaptive learning, so that if the underlying function of the network changes, the weights can adapt to the new function.

Spatio-temporal spiking network synthesis

The relative merits of rate-encoding and time- or place-encoding of neural information is a subject of frequent and ongoing debate. There are strong arguments and evidence that the mammalian neural system uses spatio-temporal coding in at least some of its systems (Van Rullen and Thorpe, 2001; Masuda and Aihara, 2003), and that this may have significant benefits in reducing energy use (Levy and Baxter, 1996) . A synthesis method which can produce networks for temporally encoded spike information will have significant benefits in terms of modeling these biological systems, and in reducing the quantity of spikes used for any given information transmission.

In this report we describe a new neural synthesis algorithm which uses the LSHDI principle to produce neurons that can implement spatio-temporal spike pattern recognition and processing; that is to say, these neurons are synthesized to respond to a particular spatio-temporal pattern of input spikes from single or multiple sources, with a particular pattern of output spikes. It is thus a method which intrinsically processes spike-time-encoded information. The synthesis method makes use of multiple synapses to create the required higher dimensionality, allowing for extreme parsimony in neurons. In most cases, only one neuron per output variable is required for most tasks, although for very complex tasks it may be more biologically realistic to break the task into a cascade of simpler sections, perhaps modeling the microstructure of a brain region. We call this method the Synaptic Kernel Inverse Method (SKIM). Training may be carried out by pseudoinverse method or any similar convex optimization, so may be online, adaptable, and biologically plausible.

This work also offers a synthesis method for networks to perform cortical sensory integration as postulated by Hopfield and Brody (2000, 2001). This required that short, sparse spatio-temporal patterns be integrated to produce recognition of a learned input. In Section 3 below, we show a detailed methodology for solving Hopfield and Brody's *mus silicium* challenge with the SKIM method.

## 2   Methods

### 2.1 The LSHDI Principle

LSHDI networks are generally represented as having three layers of neurons – the classic input, hidden and outer layer feedforward structure (see Figure 1). Should a memory function be desired, the hidden layer may have recurrent connections. However, LSHDI networks differ from regular feedforward networks in three important respects. The hidden layer is usually much larger than the input layer (at least ten times larger, as a rule of thumb). The connections from the input layer to the hidden layer are randomly generated, and are not changed during training. Finally, the output layer neurons have a linear response to their inputs.

Spatio-temporal spiking network synthesis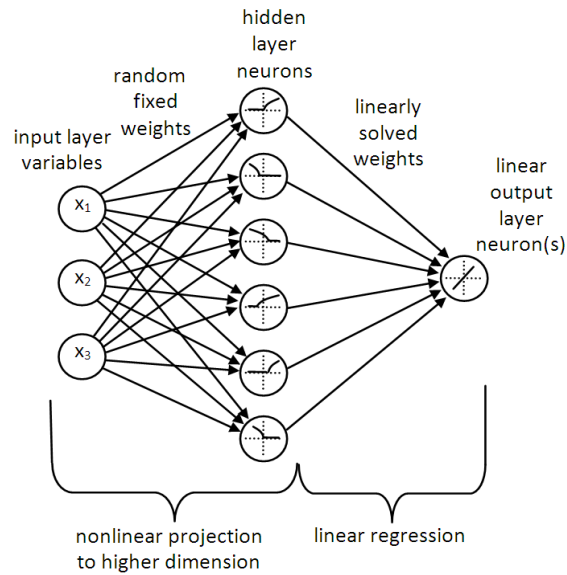

**Figure 1:** A typical LSHDI network. The input variables are projected to a higher dimension (in this case, from 3D to 6D) by means of random fixed weights and a nonlinear transformation (which in the case of NEF may be a leaky integrate-and-fire neuron, as inferred here). The outputs from the higher dimensional space are weighted and summed by linear output neurons, allowing for solution of the output weights by linear regression or classification.

The key to the success of LSHDI networks is that they embody the "kernel trick" which lies at the core of kernel methods such as kernel ridge regression and SVMs. The kernel trick is a process by which data points or classes which are not linearly separable in their current space, are projected nonlinearly into a higher dimensional space (this assumes a classification task). If the projection is successful, the data are linearly separable in the higher dimensional space. In the case of regression or function approximation tasks, the problem of finding a nonlinear relationship in the original space is transformed into the much simpler problem of finding a linear relationship in the higher dimensional space, i.e. it becomes a linear regression problem; hence the name Linear Solutions of Higher Dimensional Interlayers.

A number of researchers have shown that random nonlinear projections into the higher dimensional space work remarkably well (Rahimi and Recht, 2009; Saxe *et al.*, 2011). The NEF and ELM methods create randomly initialized static weights to connect the input layer to the hidden layer, and then use nonlinear neurons in the hidden layer (which in the case of NEF are usually leaky integrate-and-fire neurons, with a high degree of variability in their population). Many other projection options have also been successful, perhaps summed up by the title of Rahimi and Recht's paper, "Weighted Sums of Random Kitchen Sinks" (Rahimi and Recht, 2009). This paper is recommended to the reader both for its admirable readability, and the clarity with which it explains the use of random projection as a viable alternative to learning in networks. As shown by Rahimi and Recht, random nonlinear kernels can achieve the same results as random weighting of inputs to nonlinear neurons.

The linear output layer allows for easy solution of the hidden-to-output layer weights; in NEF this is computed in a single step by pseudoinversion, using singular value decomposition. In principle, any least-squares optimal regression method would work, including, for example, linear backpropagation or even biologically plausible competitive spiking networks (Afshar *et al.*, 2012). We note that for a single-layer linear backpropagation solution such as this, the problem of getting trapped in a local minimum does not occur.

The LSHDI method has the advantages of being simple, accurate, fast to train, and almost parameter-free – the only real decisions are the number of interlayer neurons and the selection of a nonlinearity, and neither of these decisions is likely to be particularly sensitive.

### 2.2 LSHDI for Spike Time Encoded Neural Representations – the SKIM Method

Spike time encoding presents difficulties for conventional neural network structures. It is intrinsically event-based and discrete rather than continuous, so networks based on smoothly continuous variables do not adapt well into this domain. Outside of simple coincidence detection, it requires the representation of time and spike history in memory (the network must remember the times and places of past spikes). The output of the network is also an event (spike) or set of events, and therefore does not map well to a linear solution space.

We have developed a biologically plausible network synthesis method in which these problems are addressed. The basic network consists of presynaptic spiking neurons which connect to a spiking output neuron, via synaptic connections to its dendritic branches, as illustrated in Figure 2. The synapses are initialized with random weights which do not change thereafter; this, together with a subsequent nonlinearity, provides the projection to a higher dimension required for the kernel trick. The dendritic branches sum the synaptic input currents. Some user-selected feature of the network – recurrent connections, axonal or dendritic delay, synaptic functions, or some combination of these - implements memory (in the form of persistence of recent spikes); and there must be a nonlinear response, which provides the nonlinearity in projection necessary for the kernel trick. In the top schematic in Figure 2 we have renamed the hidden layer as synapses, to emphasize that these (the hidden layer elements) are not spiking neurons.

Spatio-temporal spiking network synthesis

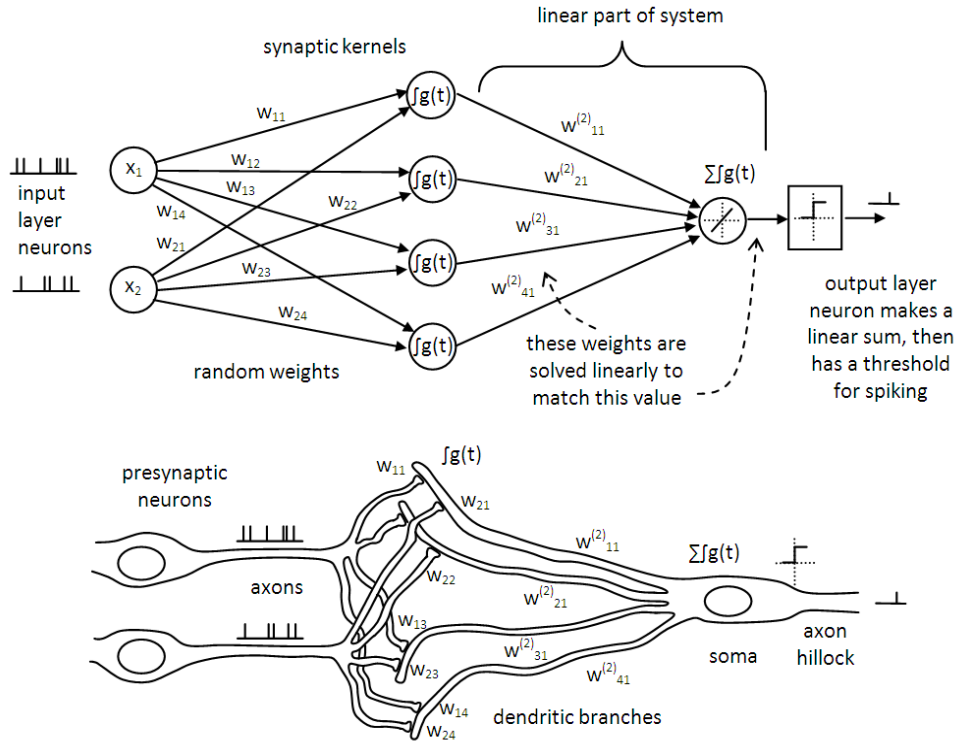

**Figure 2:** The SKIM network structure for time-encoded spike processing, shown in LSHDI and biological form. Presynaptic neurons are connected to a postsynaptic neuron through randomly generated, fixed weighted synapses. The postsynaptic dendritic branch acts as a hidden layer element, and integrates the synaptic currents with nonlinear leakage, or some equivalent nonlinearity. Memory may be implemented specifically as axonal or dendritic delays, or in terms of axonal functions. Dendritic signals are summed at the soma, and if they exceed a threshold, the axon hillock emits a spike.

The outputs from the dendritic branches are summed in the soma of the output neuron. At this stage we are able to use a linear solution to calculate the correct weights for the connection between dendritic branches and soma; solution by pseudoinverse or backpropagation will both work.

The linear solution solves the dendritic weights required to produce soma values which are below threshold for non-spike times and above threshold for spike times. The soma potential value for which the linear weights are calculated can be set to be one of two binary values, as in a classifier output; for example, it can be set to unity at spike output times, and zero when no spike is wanted. The final output stage of the neuron is a comparator with a threshold for the soma value, set at some level between the spike and no-spike output values. If the soma potential rises above the threshold, a spike is generated; and if it does not, there is no spike. This represents the generation of an action potential at the axon hillock.

The reason that this network works is that it converts discrete input events into continuous-valued signals within the dendritic tree, complete with memory (the synapses and dendritic branches may be thought of as infinite-impulse response filters); and at the same time this current and historic record of input signals is projected nonlinearly into a higher-dimensional space. The spatio-temporal series of spikes are translated into instantaneous membrane potentials. We can then solve the linear relationship between the dendritic membrane potentials and the soma potential, as though it was a time-independent classification problem: given the current membrane state, should the output neuron spike or not? The linear solution is then fed to the comparator to generate an event at the axon of the output neuron.

The inputs to this method do not necessarily need to be spikes. The method will work to respond to any spatio-temporal signals which fall within an appropriate range of magnitude. However, given that the target for this work is synthesis of spatio-temporal spike pattern processing systems, we analyze the system for spiking inputs.

### 2.3 Synaptic Kernels

In the SKIM method, the hidden layer synaptic structure performs three functions:

1. The axon signals are weighted and transmitted to the dendritic branch, which sums inputs from several axons.
2. The axon signals are nonlinearly transformed. This is necessary to ensure the nonlinear projection to a higher dimension; a linear projection would not improve the separability of the signals.
3. The axon signals are integrated, or otherwise transformed from Dirac impulses into continuous signals which persist in time, in order to provide some memory of prior spike events. For example, the use of an alpha function or damped resonance to describe the synaptic transfer of current, as is common in computational neuroscience, converts the spikes into continuous time signals with an infinite impulse response.

The sum of these transformed signals represents the proximal dendritic response to its synaptic input.

As mentioned in the previous section, steps 1, 2 and 3 may be re-ordered, given that step 3 is most likely to be linear. Any two of the steps may be combined into a single function (for example, integrating the summed inputs using an integrator with a nonlinear leak).

We refer to the hidden layer neuron structure that performs steps 1-3 above as the *synaptic kernel*. It is generally defined by the synaptic or post-synaptic function used to provide persistence of spikes, and this may be selected according to the operational or biological requirements of the synthesis. We have used leaky integration, nonlinear leaky integration, alpha functions, resonant dendrites, and alpha functions with fixed axonal or dendritic delays; all of which work to a greater or lesser extent, if their decay time is of a similar order of magnitude to the length of time for which prior spikes must be remembered. Linear leaky integration is equivalent to a recurrent network connection (with gain chosen to ensure stability). We characterized this as an infinite-impulse response filter earlier, but we note that it may be reasonable to truncate the spike response in time (or use a finite-response function) to ensure stability. Table 1 shows some typical synaptic functions in mathematical and graphical form.

# Spatio-temporal spiking network synthesis

| Kernel Type | Mathematical Expression | Typical Function (Spike at $t = 0$) |
|---|---|---|
| Stable recurrent connection (leaky integration) with nonlinear leak | $g(t) = \dfrac{1}{1+(g(t-\tau))^2} \int_{t_0}^{t} \sum_{i=1}^{L} w_{ji}^{(1)} x_{i,t} \, dt$ | 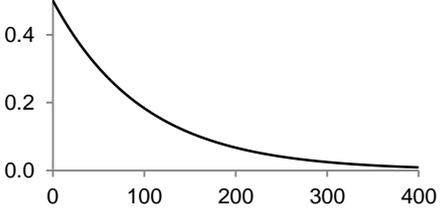 |
| Alpha function followed by compressive nonlinearity (not shown) | $g(t) = \left[ \sum_{i=1}^{L} w_{ji}^{(1)} x_{i,t} \right] \dfrac{t}{\tau} e^{-\frac{t}{\tau}}$ | 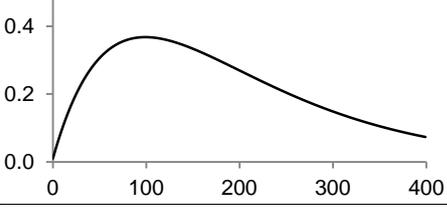 |
| Damped resonant synapse followed by compressive nonlinearity (not shown) | $g(t) = \left[ \sum_{i=1}^{L} w_{ji}^{(1)} x_{i,t} \right] e^{-\frac{t}{\tau}} \sin(\omega t)$ | 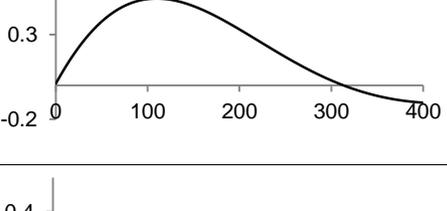 |
| Synaptic or dendritic delay with alpha function, followed by compressive nonlinearity (not shown) | for $t \geq \Delta t$:<br>$g(t) = \left[ \sum_{i=1}^{L} w_{ji}^{(1)} x_{i,t} \right] \dfrac{t-\Delta T}{\tau} e^{-\frac{t-\Delta T}{\tau}}$<br>$t < \Delta t: g(t) = 0$ | 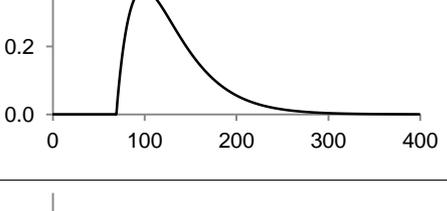 |
| Synaptic or dendritic delay with Gaussian function, followed by compressive nonlinearity (not shown) | $g(t) = \left[ \sum_{i=1}^{L} w_{ji}^{(1)} x_{i,t} \right] \dfrac{1}{\sigma\sqrt{2\pi}} e^{-\frac{(t-\Delta T)^2}{2\sigma^2}}$ | 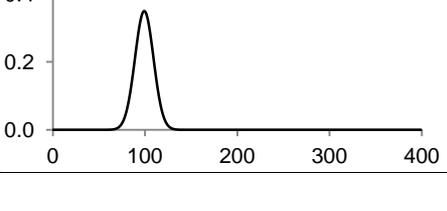 |

**Table 1:** Typical synaptic kernels in mathematical and graphical form. The order from top to bottom provides increasingly precise delay timing; with the exception of the leaky integrator, time constants were chosen for maximum synaptic transmission at 100 timesteps after spiking. Variables are as for Eq. (1). $\tau$ is the time constant for the various functions, $\Delta T$ is an explicit synaptic or dendritic delay, and $\omega$ the natural resonant frequency for a damped resonant synaptic function. The functions which do not have a compressive nonlinearity illustrated, would be followed by a standard logistical function or similar.

The synaptic kernels perform a similar synthetic function to wavelets in wavelet synthesis. By randomly distributing the time constants or time delays of the functions, a number of different (albeit not necessarily orthogonal) basis functions are created, from which the output spike train can be synthesized by linear solution to a threshold. An analogous process is spectral analysis by linear regression, in which the frequency components of a signal, which may not necessarily be orthogonal Fourier series basis functions, are determined by least-squares error minimization (Kay and Marple, 1981).

We may address the issue of resetting (hyperpolarizing) the soma potential after firing an output spike. This is simple to implement algorithmically (one can simply force all the dendritic potentials to zero after a firing event) and may improve the accuracy; our experiments with this have not shown a significant effect, but it may be present in other applications.

### 2.4 Analysis of the SKIM Method

The SKIM method may be implemented using a number of different synaptic kernels, but we can outline the method for a typical implementation. The inputs may be expressed as an ensemble of signals $\bar{x}_t \in \mathbb{R}^{L \times 1}$ where $t$ is a time or series index, and each element of $x$ represents the output of a presynaptic neuron. For convenience, the signal magnitudes may take values $x_i \in \{0,1\}$ depending on whether there is a spike from neuron $i$ at time $t$ or not. The signals propagate from the presynaptic axons to synaptic junctions with the dendritic branches of the postsynaptic neuron (for the sake of clarity, we will restrict the output to a single postsynaptic neuron at this stage; note that each dendritic branch has different characteristics, and hence has a unique index $j$). The synaptic weights $w_{ji}^{(1)}$ are fixed to random values (we can postulate a uniform distribution in some sensible range, although Rahimi and Recht (2009) have shown that this is not necessary). The superscript indicates the weights' layer. The dendrites and output neuron process the incoming signals as follows:

$$y_{n,t} = \sum_{j=1}^{M} w_{nj}^{(2)} \int_0^t g_j\left(\sum_{i=1}^{L} w_{ji}^{(1)} x_{i,t}\right) dt, \quad (1)$$

$$z_{n,t} = Boolean(y_{n,t} > \theta)$$

where $g_j(\ )$ is a nonlinear function of the weighted and summed input spikes; we may use different functions for different dendrites, hence the subscript. Note that the nonlinear function and the integral may be swapped, i.e. of the form $g(\int \sum wx\ dt)$, if that better represents the required neural functionality; the LSHDI method works in either case. Here $y_t \in \mathbb{R}^{N \times 1}$ is the output from the linear soma element, prior to thresholding; the output after comparison with threshold $\theta$ is $z_t \in \{0,1\}^{N \times 1}$ (spikes or no spikes). Each soma element $y_{n,t}$ is a linear sum of the $M$ hidden layer dendritic outputs weighted by $w_{nj}^{(2)}$. $n$ is the output (soma) vector index, $j$ the hidden layer (dendritic) index, and $i$ the input (neuron) vector index. The dendritic outputs depend on the dendrite's nonlinear function $g()$ and the randomly determined synaptic weights $w_{ji}^{(1)}$ between input and hidden layer.

As mentioned previously, the synaptic weights $w_{ji}^{(1)}$ are randomly set (usually with a uniform distribution in some appropriate range) and remain fixed. The training of the

Spatio-temporal spiking network synthesis network consists of calculating the weights $w_{nj}^{(2)}$ connecting the dendrites to the soma. This is performed in a single step (for a batch learning process) using a linear regression solution. If we define the dendrite potentials at the synapses to be

$$a_{j,t} = \int_0^t g_j\left(\sum_{i=1}^L w_{ji}^{(1)} x_{i,t}\right) dt, \tag{2}$$

we can represent the outputs $a$ of the hidden layer in the form of a matrix $A$ in which each column contains the hidden layer output for one sample in the time series, with the last column containing the most recent sample; $A = [a_1 \cdots a_k]$ where $A \in \mathbb{R}^{M \times k}$. Similarly we can construct a matrix $Y$ of the corresponding output values; $Y = [y_1 \cdots y_k]$ where $Y \in \{0,1\}^{N \times k}$. Note that $Y$ is expected to consist of binary or Boolean values; spike or non-spike. Synthesizing the network requires that we find the set of weights $W \in \mathbb{R}^{N \times M}$ that will minimize the error in:

$$WA = Y. \tag{3}$$

This may be solved analytically by taking the Moore-Penrose pseudoinverse $A^+ \in \mathbb{R}^{k \times M}$ of $A$:

$$W = YA^+. \tag{4}$$

In a batch process, $A$ and $Y$ will be static data sets and the solution can be obtained by means of singular value decomposition (SVD). In a recent report (Tapson and van Schaik, 2013), we have described an incremental method for solving the pseudoinverse, which we called OPIUM – the Online PseudoInverse Update Method. OPIUM may be used for online learning, or where batch data sets are so large that the matrix sizes required for the SVD are too unwieldy in terms of computational power and memory.

The function used to provide persistence of spikes may be selected according to the operational or biological requirements of the synthesis. We have used a number of mathematically definable nonlinearities, but there is no reason why others, including arbitrary functions that may be specified by means of e.g. a lookup table, could not be used. There is no requirement of monotonicity, and we have successfully used wavelet kernels such as the Daubechies function, which are not monotonic.

## 3 Results

### 3.1 An Example of the SKIM Method

Consider a situation in which we wish to synthesize a spiking neural network that has inputs from four presynaptic neurons, and emits a spike when, and only when, a particular spatio-temporal pattern of spikes is produced by the pre-synaptic neurons. We create an output neuron with 80 dendritic branches, and make a single synapse between each presynaptic neuron and each dendritic branch, for a total of 320 synapses. (This gives a "fan-out" factor of 20 dendrites per input neuron, which is an arbitrary starting point; we will discuss some strategies for reducing synapse and dendrite numbers, should synaptic parsimony be a goal). The structure is therefore

Spatio-temporal spiking network synthesis four input neurons, each making one synapse to each of eighty dendritic branches of a single output neuron.

The pattern to be detected consists of four spikes, one from each neuron, separated by specific delays. This pattern will be hidden within random "noise" spikes (implemented with a Poisson distribution) – see Fig. 3.

In this example, we use the following functions for summing, nonlinearity, and persistence. A summed signal $u_{j,t}$ is obtained conventionally:

$$u_{j,t} = \sum_{i=1}^{L} w_{ji}^{(1)} x_{i,t} \qquad (5)$$

Note that $L=4$ in this example, and the weights $w_{ji}^{(1)}$ are randomly (uniformly) distributed in the range (-0.5, 0.5). After summing, the logistical function is used to nonlinearly transform the summed values:

$$v_{j,t} = \frac{1}{1+e^{-ku_{j,t}}} - 0.5 \qquad (6)$$

Here $k=5$ is a scaling constant.

The alpha synaptic function is used to provide persistence of the signal in time, as follows: at any timestep $t_0$, if $v_{j,t_0} \neq 0$ (i.e. there is a spike), a function $z_{j,t}(v_{j,t_0})$, $t > t_0$, is added to the dendritic branch signal $a_{j,t}$. $z_{j,t}$ is an alpha function scaled by the amplitude of $v_{j,t_0}$ and with its origin at $t_0$:

$$z_{j,t} = v_{j,t_0} \left( \frac{t-t_0}{t_s} e^{-\frac{t-t_0}{t_s}} \right) \qquad (7)$$

The time constant $t_s$ of the alpha function will define the persistence of the spike input in time. In practice we have found $t_s \approx t_{max}/2$ to be a useful heuristic, where $t_{max}$ is the longest time interval for which spikes will need to be "remembered". In this case, the maximum length of the pattern was 200 timesteps, and the values of $t_s$ were uniformly distributed in the range (0, 100) timesteps, thereby straddling the heuristic value. This heuristic applies only to the alpha function; other kernels will require some random distribution of time constants or delays in some similarly sensible range.

Figure 3 shows the development of the signals through the system.

Spatio-temporal spiking network synthesis

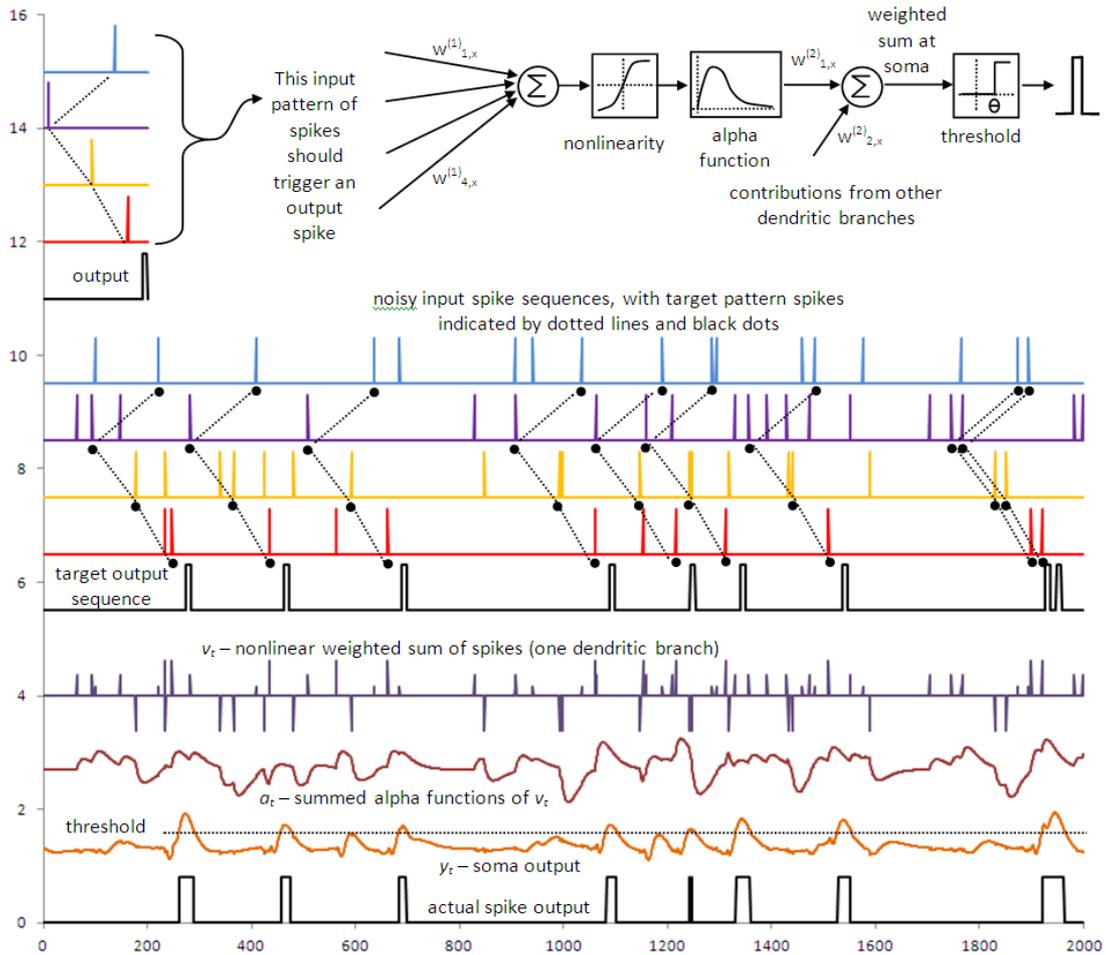

**Figure 3**: This shows the development of the SKIM method for a spatio-temporal spike pattern recognition system. The structure of one dendritic branch and the soma is shown at the top. The signals are, from top: the input pattern on four channels and the target output; a test sequence with added noise spikes, and target output; the nonlinear summed spike output from one dendritic branch; the resulting alpha function output from that branch; the soma potential for the output neuron; and the resulting output spike train. It can be seen that the spike pattern is successfully recognized in the presence of some spike noise.

The network was presented with a mixture of Poisson-distributed random spikes and Poisson-distributed spike patterns, such that the number of random noise spikes was approximately equal to the number of pattern spikes. An output spike train was used to provide the solution data during the training calculation (note that while input events were effectively instantaneous, taking just one time step, the output train consisted of slightly wider windows of ten time steps, on the basis that a spike within some short window would constitute a functional output; the output spike is also displaced moderately later in time than the last input spike, in order to allow for the dendritic response to peak). The broader target "spike" causes the network to train for a broad output spike, the effect of which may be seen more clearly in Figure 5. The results for the present synthesis can be seen in Figure 3.

Spatio-temporal spiking network synthesis

In examining the issue of resetting the somatic potential, we note that it is generally accepted that the Markov property applies to integrate-and-fire or threshold-firing neurons (Tapson *et al*., 2009); so that the dependence of the firing moment of a neuron is not dependent on the history of the neuron prior to the most recent spike. It might seem intuitively necessary that this only holds if the neuron is reset to a potential of zero (hyperpolarized) after the most recent spike, but in fact from the point of view of the trajectory of the membrane potential, and the inverse solution of the dendritic weights needed to produce that trajectory, it is immaterial what the potential starting level is, as long as it is defined and consistent. Those who are concerned by this issue may cause their simulation code to reset the membrane potential after spiking.

### 3.2 Use of the SKIM Method on a Predefined Problem

In this section we will illustrate the use of the SKIM method to solve a problem in spatio-temporal pattern recognition. In 2001, John Hopfield and Carlos Brody proposed a competition around the concept of short-term sensory integration (Hopfield and Brody, 2000; Hopfield and Brody, 2001). Their purpose was to illustrate the usefulness of small networks of laterally- and recurrently-connected neurons, and part of the competition was to develop a network to identify words based on a very sparse representation of audio data. The words were spoken digits drawn from the TI46 corpus (TI46, 2013), and were processed in a quasi-biological way; they were passed through a cochlea-like filterbank to produce 20 parallel narrowband signals, and then the times of onset, offset and peak power were encoded as single spikes at that time, in separate channels; so, each word was encoded as an ensemble of single spikes on multiple channels.

Hopfield and Brody's neural solution – referred to as *mus silicium*, a mythical silicon-based mouse-like lifeform – was based on neurons which exhibited bursting spiking, with a linearly decaying time response, to input spikes. The key to its operation was that this linear decay offered a linear conversion of time to membrane potential amplitude, and thereby the encoding of time, which then enabled the recognition of spatio-temporal patterns; and that coincidence of signal levels could be detected by synchronized output spiking. In the SKIM method, we achieve a similar result (without bursting spikes), using synapses with arbitrary time responses, which allows a significantly greater degree of biological realism together with sparser use of neurons and sparser use of spikes. It remains to be shown that the SKIM method is actually capable of solving the problem, and we outline its use for this purpose here.

Hopfield and Brody pre-processed the TI46 spoken digits to produce 40 channels with maximally sparse time encoding – a single spike, or no spike, per channel per utterance (a full set of onset, offset and peak for all 20 narrowband filters would require 60 channels, but Hopfield and Brody chose to extract a subset of events – onsets in 13 bands, peaks in 10 bands, and offsets in 17 bands). The spikes encode onset time, or peak energy time, or offset time for each utterance. Examples are shown in Figure 5.

Hopfield and Brody's original *mus silicium* network contained three or four layers of neurons, with an input layer (arguably two layers, as it spreads the input from 40 to 800 channels); a hidden layer with excitatory and inhibitory neurons, and significant numbers of lateral connections (75-200 synaptic connections each); and an output layer with one neuron per target pattern. The input layer was not encoded as one input per channel, but each channel was encoded with 20 different delays, to produce 800 input neurons. There were apparently 650 hidden layer neurons (a number of 800 is also referred to; elsewhere, *mus silicium* is describfed as having 1000 neurons in total). Input layer neurons were designed to output a burst of spikes for each input spike, so the original input pattern of 40 spikes would be scaled up to 800 series of 20-50 spikes each, as the inputs to the hidden layer.

By contrast, we will demonstrate the use of the SKIM method to produce a feedforward –only network with just two layers of neurons – 40 input neurons (one per input channel) and 10 output neurons (one per target pattern). The presynaptic neurons will be connected by ten synapses each to each postsynaptic neuron, for a total of 400 synapses per postsynaptic (output) neuron. This gives the network a total of 50 spiking neurons connected by 4000 synapses.

The exact choice of synaptic kernel is not critical for success in this system. A simple α-function performs extremely well, as do synapses with a damped resonant response. In the data which follow, we show results for a number of different functions.

The prescribed training method for *mus silicium* was extremely stringent; it could be trained on only one single utterance of the target digit ("one"), interspersed with nine randomly selected utterances of other digits. The task was a real test of the ability of a network to generalize from a single case. In order to achieve the robustness to time-warping of the utterances for different speakers and different speech cadences, we produced a training set in which the exemplar pattern and its nine random companions were reproduced with a range of time warping from 76% to 124% of the originals.

Having been trained on this very small data set, the network is then tested on the full set of 500 utterances (which includes the examplar and nine random utterances, and therefore has 490 unseen utterances), almost all by previously unheard speakers.

## 4   Results

There are no published data for the accuracy of Hopfield and Brody's network, but the winning entry in their competition, from Sebastian Wills, is extensively described (Wills, 2001; Wills, 2004). The network was tested with 500 utterances of the digits 0-9, giving 50 target utterances of the digit "one" (only one of which was the exemplar) and 450 non-targets; and the error was defined as:

$$error = \frac{\#\,false\,negatives}{\#\,true\,positives} + \frac{\#\,false\,positives}{\#\,true\,negatives} \qquad (8)$$

Wills' minimum error was 0.253. Errors smaller than this are easy to achieve with SKIM – see Table 2 below. Figure 4 shows performance on the test data set for the neuron trained on the single training utterance of "one", and is the equivalent for a SKIM network to Wills' Figure 2.18 (Wills, 2004: p.26).

Spatio-temporal spiking network synthesis

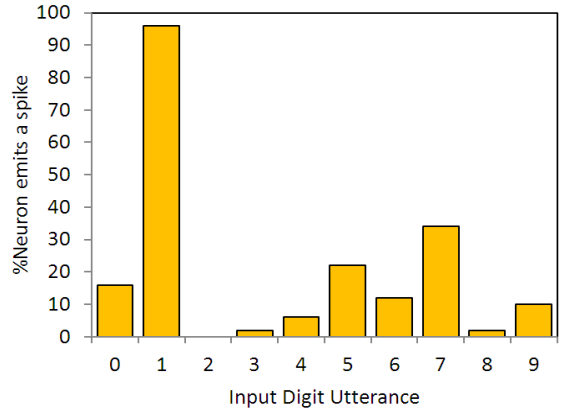

**Figure 4**: Output neuron responses for a neuron synthesized to spike in response to the utterance of digit "one", trained on a single exemplar. The data show responses for the full set of 500 previously unseen time-encoded digit utterances, as in (Hopfield and Brody, 2000, 2001; Wills, 2001, 2004).

| Network | Error |
|---|---|
| Wills, 2001 | 0.253 |
| SKIM, Alpha synapse | 0.224 |
| SKIM, Damped resonance | 0.183 |
| SKIM, Delay plus alpha | 0.173 |
| SKIM, Delay plus Gaussian | 0.169 |

**Table 2:** Errors for SKIM networks applied to the *mus silicium* problem, with various different types of synaptic kernels. All networks had the minimum 40 input neurons and 10 output neurons, and were connected with 40x10x10 = 4000 synapses in total.

The authors would like to make it clear that the results in Table 2 do not imply that this network would have won the *mus silicium* competition, as that competition had explicit restrictions on synaptic time constants that would have excluded a SKIM network; there was also a requirement for a test of robustness to weight change, that is not practically applicable in a network with two layers of neurons (the competition assumed a three-layer network; or strictly speaking, four layers of neurons if the initial input spreading is taken into account). Nonetheless, we believe the SKIM performance on this problem illustrates its usefulness as a synthesis method for spatio-temporal pattern recognition.

Figure 5 illustrates some spike raster patterns for this application.

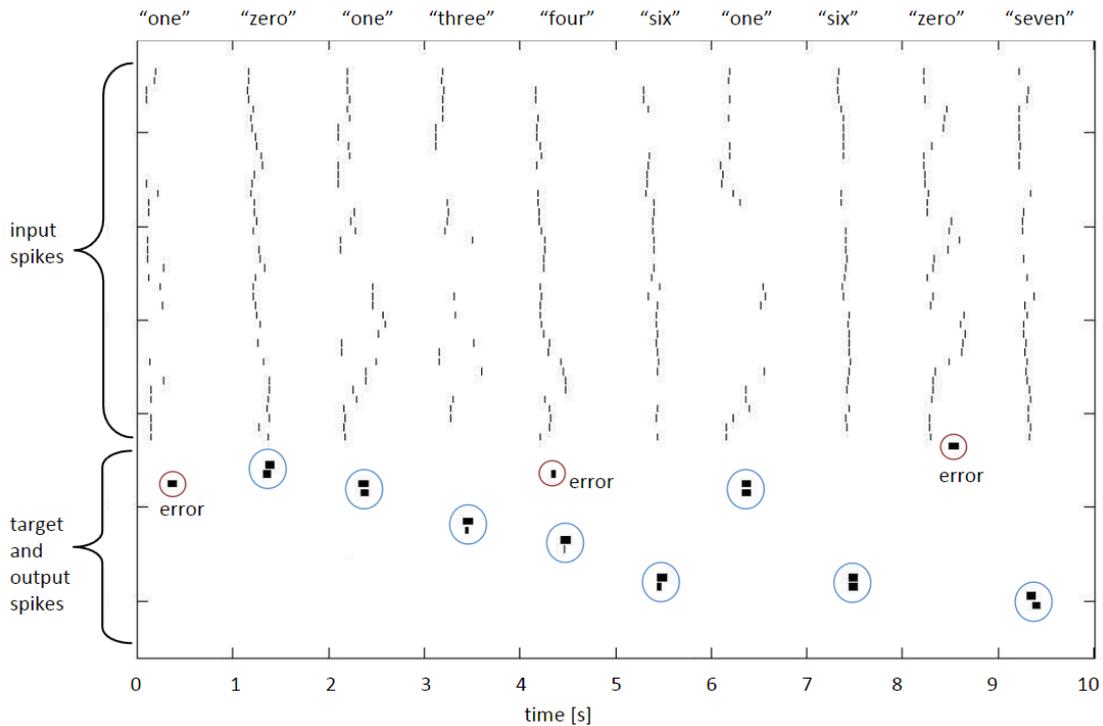

**Figure 5:** Spike rasters for ten spoken digits, showing input, target and output spikes. The spike pairs circled with blue dotted lines indicate correct classifications (target and output spikes have been placed together in the raster plot to make visual assessment easier); the other output spikes, circled in red, are errors. The breadth of the target and output spikes – approximately 20 standard spike intervals - is explained in the text below.

### 4.1 Target Spike Implementation

The *mus silicium* competition data set illustrates an interesting question for synthesis of networks with spiking output: what and when should the output be? If we adhere to the spiking paradigm, then the output should be a spike, but at what time, relative to the input spike set? The TI46 digits are nominally situated in one-second-long windows, but the variations in spike onset and offset times show that this is by no means a consistent or reliable centering. We arbitrarily used the time of the last spike of the training set digits as the reference time, within each 1-second window, when the output spike should occur. A glance at Figure 5 shows that when this is applied to the testing patterns, the "target" spike has often commenced before all the input spikes have occurred, which is obviously sub-optimal from a detection perspective. We made a poor compromise in this case, by spreading the energy of the output spike over 200 ms (hence the visible length in Fig. 5), so that there was in effect a lengthy output or target window during which the spike would occur. A moment's thought by the reader will suggest several different and possibly better ways in which this might be done; nonetheless the network shows useful results with this method, and further research will no doubt improve the performance.

A feature of the pseudoinverse solution method is that it is sensitive to the energy in the target signal, so that a target spike with a nominal amplitude of 10 units and duration of one time step enforces a more significant learning response than a target spike with an amplitude of one unit and the same duration. A single Dirac delta spike, occurring in a period of say a thousand time steps where the output is otherwise zero, provides very little incentive for the learning process to move the solution away from a zero output, and so we suggest that increasing the spike amplitudes above unit level may improve the accuracy of this method.

### 4.2 Errors and Capacity

Whilst the SKIM method manages to avoid the large number of spiking neurons used in NEF synthesis, it might be argued that the number of synapses is still unrealistically large in comparison with the complexity of the problem, and that we have replaced the profligate use of spiking neurons with a profligate use of synapses. Current estimates suggest there are on average 7000 synapses per cortical neuron in the adult human brain (Drachman, 2004), so it's not immediately obvious what a correct proportion of synapses might be. We note that biological and computational evidence supports ongoing synaptic pruning as critical in brain function (Paolicelli *et al.*, 2011) and dynamic network optimization (Checik *et al.*, 1999), so we present here some strategies for reducing synaptic numbers by strategic pruning.

In Figure 6 we show the weights for the 80 dendrites in the problem of Figure 4 (for ten different iterations, i.e. ten different nonlinear random projections). It can be seen that approximately 25% of the dendrites are contributing 50% of the weight of the solution; 50% of the dendrites contribute 80% of the weight of the solution. We can follow the physiological practice and prune the dendrites or synapses that are not contributing significantly to the solution. Note that the linear weights must be re-solved after synapses are pruned, or the solution will be non-optimal.

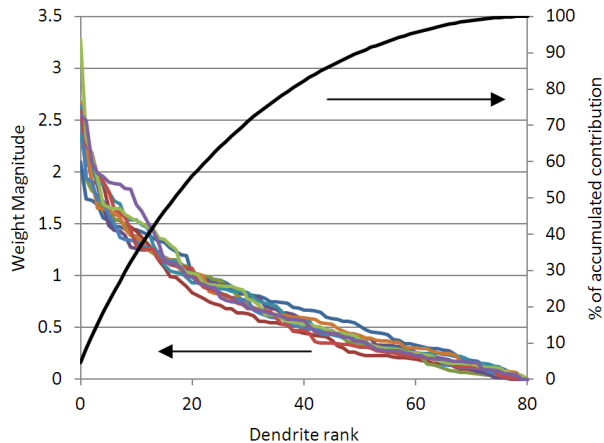

**Figure 6:** The magnitude of the solved dendritic weights for 80 dendrites, in ten different solutions of the example problem of Fig. 3, are shown (left axis). It can be seen that in all cases, the 20 largest weighted dendrites are contributing over 50% of the solution magnitude; and that the 40 largest weights are contributing over 80% of the solution (right axis). This suggests that pruning the lowest weighted dendrites will not significantly alter the accuracy of the solution.

There are numerous strategies by which the weights can be pruned. Two strategies which we have used with success are to over-specify the number of synapses and then prune, in a two-pass process; or to iteratively discard and re-specify synapses. For example, if we desire only 100 synapses, we can synthesize a network with 1000 synapses; train it; discard the 900 synapses which have the lowest dendritic weights associated with them; and then re-solve the network for the 100 synapses which are left. This is the two-pass process. Alternatively, we can specify a network with 100 synapses; train it; discard the 50 synapses with the lowest weights, and generate 50 new random synapses; re-train it; and so on – this is the iterative process. The choice of process will depend on the computational power and memory available, but both of these processes produce networks which are more optimal than the first-order network produced by the SKIM method.

## 5  Discussion

The SKIM method offers a simple process for synthesis of spiking neural networks which are sensitive to single and multiple spikes in spatio-temporal patterns. It produces output neurons which may produce a single spike or event, in response to recognized patterns on a multiplicity of input channels. The number of neurons is as sparse as may be required; in the examples presented here, a single input neuron per channel, representing the source of input spikes, and a single output neuron per channel, representing the source of output spikes, has been used. The method makes use of synaptic characteristics to provide both persistence in time, for memory, and the necessary nonlinearities to ensure increased dimensionality prior to linear solution. The learning method is by analytical pseudoinverse solution, so has no training parameters, and achieves optimal solution with a single pass of each sample set. We believe that this method offers significant benefits as a basis for the synthesis of all spiking neural networks which perform spatio-temporal pattern recognition and processing.

**Conflict of Interest**

This research was conducted in the absence of any commercial or financial relationships that could be construed as a conflict of interest.

**Acknowledgements**

The authors thank James Wright for help with data preparation, and the organizers of the CapoCaccia and Telluride Cognitive Neuromorphic Engineering Workshops, where these ideas were formulated.

Spatio-temporal spiking network synthesis